\documentclass[journal]{IEEEtran}
\ifCLASSINFOpdf
\usepackage[pdftex]{graphicx}
\graphicspath{{imgs/}}
\DeclareGraphicsExtensions{.pdf,.jpeg,.png}
\else
\fi
\usepackage{amsmath}
\usepackage{algorithm}
\usepackage{algorithmic}
\usepackage{array}
\newcolumntype{C}[1]{>{\centering\let\newline\\\arraybackslash\hspace{0pt}}m{#1}}
\usepackage{booktabs}

\usepackage{url}
\usepackage{multirow}
\usepackage{rotating}
\usepackage{mathtools}

\begin{document}
\title{A Century of Portraits:\\
A Visual Historical Record\\
of American High School Yearbooks\textsuperscript{\small{\dag}}}

\author{Shiry~Ginosar, Kate~Rakelly, Sarah~M.~Sachs, Brian~Yin, Crystal~Lee, Philipp~Kr{\"a}henb{\"u}hl and~Alexei~A.~Efros%
\thanks{\dag This is an extended version of a paper that first appeared as~\cite{Ginosar_2015_ICCV_Workshops}. Sections IV.B.2-3 and VI contain new material while section V has been expanded.}%
\thanks{S. Ginosar, K. Rakelly, B. Yin, C. Lee and~A.~A. Efros are with the department of Electrical Engineering and Computer Science, UC Berkeley.}%
\thanks{P. Kr{\"a}henb{\"u}hl is with the Department of Computer Science, UT Austin and was with the Department of Electrical Engineering and Computer Science, UC Berkeley during the majority of this work.}%
\thanks{S.~M. Sachs was with Brown University.}}%

\maketitle

\begin{abstract}
Imagery offers a rich description of our world and communicates a volume and type of information that cannot be captured by text alone.
Since the invention of the camera, an ever-increasing number of photographs document our ``visual culture" complementing historical texts.
But currently, this treasure trove of knowledge can only be analyzed manually by historians, and only at small scale.
In this paper we perform automated analysis on a large-scale historical image dataset.
Our main contributions are:
1) A publicly-available dataset of 168,055 (37,921 frontal-facing) American high school yearbook portraits.
2) Weakly-supervised data-driven techniques to discover historical visual trends in fashion and identify date-specific visual patterns.
3) A classifier to predict when a portrait was taken, with median error of 4 years for women and 6 for men.
4) A new method for discovering and displaying the visual elements used by the CNN-based date-prediction model to date portraits, finding that they correspond to the tell-tale fashions of each era.
\end{abstract}


\IEEEpeerreviewmaketitle

\section{Introduction}
\IEEEPARstart{I}{n} their quest to understand the past, historians---from Herodotus to the present day---primarily rely on textual records.
However, some details are perceived as too mundane to put down in writing or too difficult to accurately describe.
For example, it would be hard for a future historian to understand what the term ``hipster glasses" refers to, just as it is difficult for us to imagine what ``flapper galoshes'' might look like from a written description alone~\cite{flappers}.
The invention of the \textit{daguerreotype} in 1839 as a means of relatively cheap, automatic image capture heralded a new age of massive visual data creation with potentially profound implications for historians.
This new format was complementary to historical texts, as it could capture those nuances and transmit non-verbal information that would otherwise be lost.

The study of history is often an exercise in finding patterns in large amounts of data.
For written accounts, historians have begun to use digital humanities techniques to automatically mine large text corpora.
For example, using Google Books it is possible to study a diverse set of topics such as word usage over time and the histories of events like the Civil War or the spread of influenza~\cite{googlebooks}.
In contrast, despite the abundance of historical visual data over the last century and a half, historians are still limited by the speed of manual curation.
There are perhaps many unseen visual connections that are missed because tools for large-scale visual data mining have yet to be introduced into the field.

\begin{figure}[t]
\begin{center}
\includegraphics[width=\linewidth]{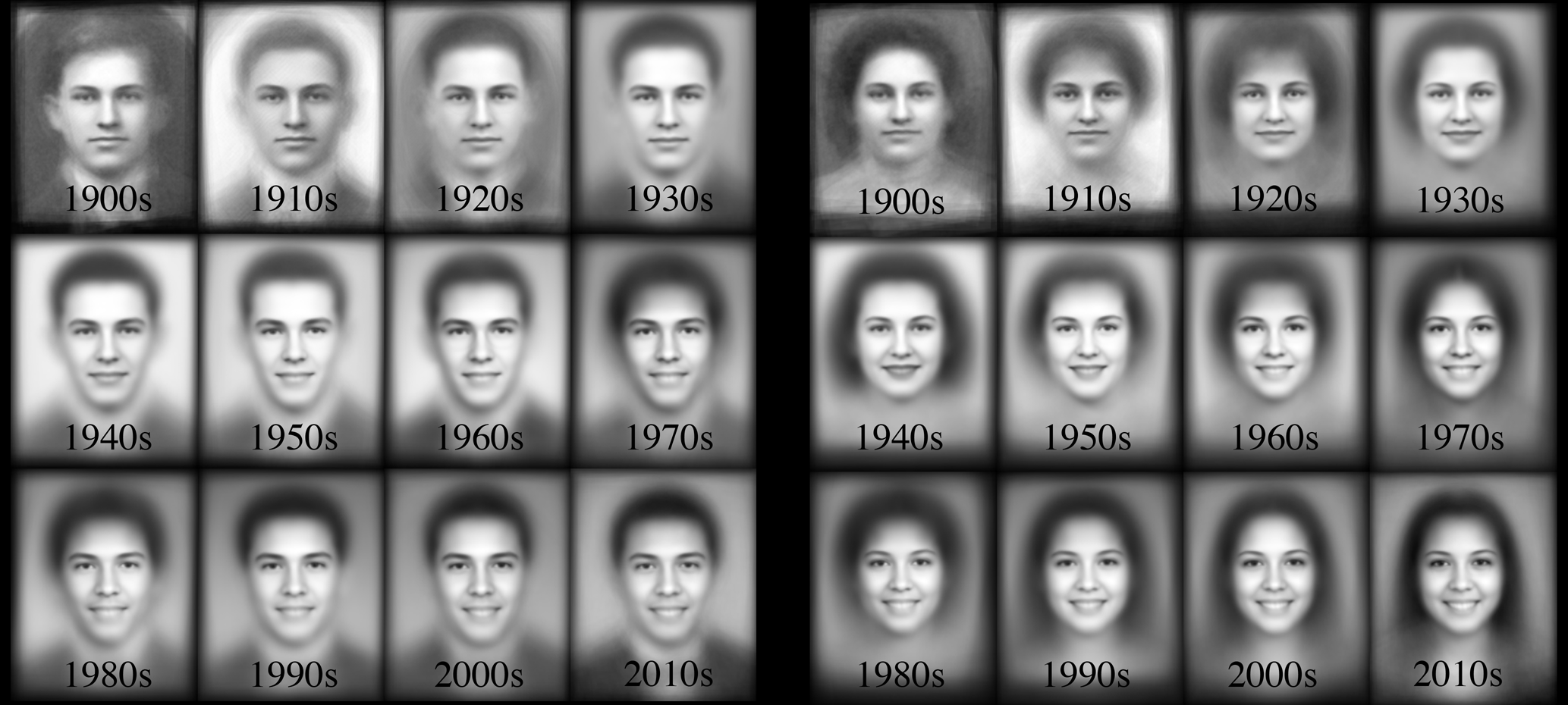}
\end{center}
   \caption{Average images of high school students by decade. The evolving fashions and facial expression throughout the 20th century are evident in this simple aggregation. For example, notice the increasing extent of smiles over the years and the recent tendency for women to wear their hair long. In contrast, note that the suit is the default dress code for men throughout.}
\label{fig:decade_averages_teaser}
\end{figure}
 
We take a new approach to the analysis of visual historical data by introducing data-driven methods suited to mining large image collections.
Specifically, we apply these methods to discovering one of the most interesting historical trends---the evolution in the appearance of people over time.
We present a collection of one type of widely available yet little used historical visual data---a century's worth of United States high school yearbook portraits (Fig~\ref{fig:decade_averages_teaser}).
Yearbooks, an iconic American high school staple, have been published since the wide adoption of film (the first Kodak camera was released in 1888) and contain standardized portrait photos of the graduating class.
Yearbook portraits provide a consistent visual format through which one can examine changes in content from personal style choices to developing social norms.
In this paper, we present a large-scale dataset of yearbook portraits spanning the entire 20th century, and report on a number of experiments to analyze it. 

First, we mine the portrait data to discover trends over time and date-specific visual patterns. We examine changes in social norms by studying the practice of smiling to the camera and men's changing hair styles during the social changes of the 1960s.
Additionally, we discover that fluctuations in the popularity of eyewear is correlated with advances in contact lens technology.
Finally, we mine for the quintessential ``look" of each decade by employing a technique of discriminative clustering.
Our data-driven results are consistent with existing historical records of the fashion trends in hair, makeup and eyewear from the 20th century.

Second, we use the time-correlated visual variability in the portraits to predict, from an image of a face alone, when the photograph was taken.
Using a convolutional neural network (CNN) trained on our dataset, we are able to date yearbook portraits within a median error of four years of their true date.
We further demonstrate some generalization to an unseen dataset of historical celebrity portraits despite the large differences in appearance between high school students and adult actresses and models.

Finally, while CNN classifiers have proven to be the leading tool for many image domains, it remains challenging to tell \textit{why} a specific classification decision has been made.
This is particularly important for tasks like dating where the labels are weak, the visual space is huge, and much of the visual data might be irrelevant to the task.  
We propose a method to discover which parts of the image were most useful for pinpointing the date in which it was taken. 
At the core of our approach lies the insight that we can disable parts of the network without altering the dating decision.

The main contributions of this paper are:
1) A publicly-available historical image dataset that comprises a large scale collection of yearbook portraiture from the last 120 years in the United States.
2) Data-driven methods to discover historical visual patterns in fashion and social norms.
3) A CNN classifier to predict the date in which a portrait was taken, with median error of 4 years for women and 6 for men.
4) A method for visualizing the time-specific elements used by the CNN to date the portraits.

\section{Related Work}

\subsubsection{Historical Data Analysis}
Researchers in the humanities tease out historical information from ever larger text corpora thanks to advances in natural language processing and information retrieval.
For example, these advances (together with the availability of large-scale storage and OCR technology) enabled Michel et al.~\cite{googlebooks} to conduct a thorough study of about 4\% of all books ever printed resulting in a quantitative analysis of cultural and linguistic trends.

To date, the automated analysis of historical images has been relatively limited.
Some examples include modeling the evolution of automobile design~\cite{cars} and architecture~\cite{linking2015iccp} as well as \textit{image dating}--determining the date when historical color photographs were taken~\cite{dating,dating2}.
Here we extend upon these works by presenting a yearbook dataset that we use to answer a broader set of questions.
Concurrent and independent of our work,~\cite{other_yearbooks} also proposed using yearbook data for image dating but focused on yearbooks from two counties in Missouri.
Our work differs in that we mine for various patterns in yearbook data beyond date prediction.
Moreover, our dataset is unique as it a broader sample of locations across the United States as well as constant coverage over time (see Figure~\ref{fig:facesperyear}).

\subsubsection{Modeling Style}
Recently several researchers began modeling fashion.
In HipsterWars, Kiapour et al.~\cite{HipsterWarsECCV14} take a supervised approach and use an online game to crowd-source human annotations that are then used to train models for style classification.
Hidayati et al.~\cite{NYCfashion} take a weakly-supervised approach to discover the recent (2010-2014) trends in the New York City fashion week catwalk shows.
They extract color and texture features and use these to discover the representative visual style elements of each season via discriminative clustering~\cite{doersch2012what}.
While we also deal with fashion and style in this paper, our focus is on changes in style through a much longer period.
Because our dataset includes scanned images from earlier time periods, much of it consists of grayscale photographs and of lower resolution than the recent datasets described above.
This makes some of the above approaches such as the usage of color and texture features unsuitable for our data.

\subsubsection{Deep Neural Networks}
Of the many CNN architectures designed in recent years, the VGG~\cite{vgg} network is one of the best-performing and most versatile.
It is designed as a deep network of $16$ convolutional layers with spatially-grouped feature maps and two fully connected layers on top.
The VGG model trained on ILSVRC 2012~\cite{ILSVRC15} has been able to generalize well to various computer vision tasks with proper fine-tuning (further training) on the target data and task.
In this paper, we use VGG for the task of portrait dating and visualize which image regions it uses to make inference decisions.

\subsubsection{Deep Neural Network Visualization}
Several attempts have been made to visually understand the inner-workings of deep networks.
One approach taken by~\cite{mahendran15understanding,DosovitskiyB15,yosinski} visualizes images that produce a specific set of features in CNNs.
Another approach aims to find input images that maximize the activation of single units in the network~\cite{yosinski,visualization_techreport}.
In the realm of faces,~\cite{torre} synthetically generate images that maximally activate individual neurons.
Unlike our method, these approaches do not explain which spatial locations in an input image contribute to the classification.

Zeiler et al.~\cite{ZeilerF14} examine which parts of the image result in the highest response of single spatial units by systematically obstructing parts the image.
They use deconvolutional networks to invert the effect of pooling layers and reconstruct an approximation of the input pixels from the activations of intermediate layers of the network.
Unlike this approach, our method outputs pixel locations rather than an approximation of the input.
Following a similar approach, Zhou et al.~\cite{scenecnn_iclr15} ask which segments of an image are most responsible for a particular classification decision.
In contrast, we do not force our visual elements to be enclosed in image regions, allowing us to discover ephemeral visual structures beyond objects. 

Most similar to our approach, Simonyan et al.~\cite{Simonyan14a} use the network gradient propagated back to pixel space for a single input image as an approximation of which spatial locations would maximize the classification score if changed.
This method discovers the spatial locations that affect the class score for a canonical image from this class and only reveals the general location of the object in the image.
In contrast, our approach takes into account the unique path which the input image takes through the network and therefore discovers which visual elements were used by the CNN to classify \textit{this} image.
As a result, our method focuses on localized areas that correspond to discriminative visual features.

\section{The Yearbook Dataset}

\begin{figure}
\begin{center}
\includegraphics[width=\linewidth]{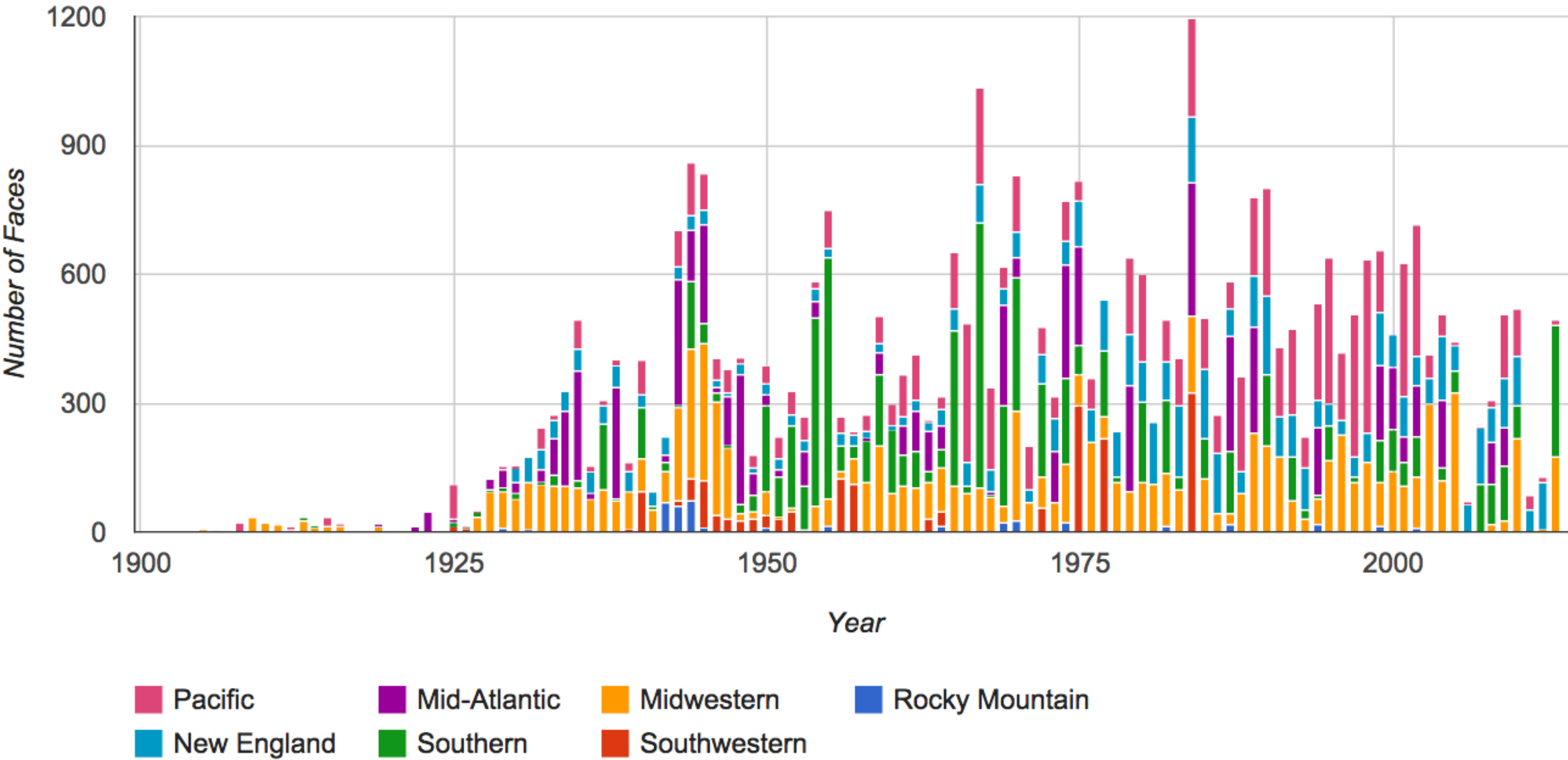}
\end{center}
   \caption{The distribution of portraits per year and region. Our dataset is unique in that it is diverse in terms of both geographic location and time coverage.}
\label{fig:facesperyear}
\end{figure}

We are at an auspicious moment for collecting historical yearbooks as it has become standard in recent years for local libraries to digitally scan their yearbook archives.
This trend enabled us to download publicly available yearbooks from various online resources such as the Internet Archive and numerous local library websites.
We collected 949 scanned yearbooks from American high schools ranging from 1905-2013 across 128 schools in 27 states.
These contain 168,055 individual senior-class portrait photographs in total along with many more underclassmen portraits that were not used in this project.
After removing all non-frontal facing we were left with a dataset of 37,921 photographs that depict individuals from 814 yearbooks across 115 high schools in 26 states.
 
On average, 28.8 faces are included in the dataset from each yearbook with an average of 329 faces per school across all years.
The distribution of photographs over year and region is depicted in Figure~\ref{fig:facesperyear}.
Overall, 46.4\% of the photos come from the 100 largest cities according to US census~\cite{100Cities}.
 
Let us consider the potential biases in our data sample as compared to the high school age population of the United States.
Since 1902 America's high schools have followed a standard format in terms of the population they served~\cite{Goldin}.
Yet, this does not mean that the population of high school students has always been an unbiased sample of the US youth population.
In the early 1900s, less than 10\% of all American 18-year-olds graduated from high school, but by end of the 1960s graduation rates increased to almost 50\%~\cite{Goldin}.
Moreover, the standardization of high schools in the United States left out most of the African American population, especially in the South, until the middle of the 20th century~\cite{GoldinRace}.

In our dataset 53.4\% of the photos are of women, and 46.6\% are of men. As the true gender proportion in the population is only available in a census year we are unsure if this is a bias in our data. However, the gender imbalance may be due to the fact that historically girls are disproportionately more likely than boys to attend high school through graduation~\cite{Goldin}.

\label{sec:data_prep}
In order to turn raw yearbooks into an image dataset we performed several pre-processing operations.
First, we manually identified the scanned pages of senior-class portraits.
After converting these to grayscale for consistency across years, we automatically detected and cropped to faces.
We then extracted facial landmarks from each face and estimated its pose with respect to the camera using the IntraFace system~\cite{intraface, intraface_final}.
This allowed us to filter out images of students who were not facing forward.
Next, we aligned all faces to the mean shape using an affine transform based on the computed facial landmarks. 
Finally, we divided the photos into those depicting males and females using an SVM in the whitened HOG feature space~\cite{Dalal05histogramsof,BharathECCV2012} and resolved difficult cases (confidence score lower than 90\%) by crowdsourcing a gender classification task on Mechanical Turk.
Our final dataset consists of cropped portraits with year, state, city, school and gender annotations.

\section{Mining the Visual Historical Record}

We demonstrate the use of our historical dataset in answering questions of historical and social relevance.
\vspace{-0.1in}
\subsection{Getting a Sense of Each Decade}
The simplest visual-data summarization technique of facial composites dates back to the 1870s and is attributed to Sir Francis Galton~\cite{galton}.
Here we use this technique to organize the portraits chronologically.
Figure~\ref{fig:decade_averages_teaser} (first page) displays the pixel-mean of images of male and female students for each decade in our data.
These average images showcase the main modes of the popular fashions in each time period.

\vspace{-0.1in}
\subsection{Capturing Trends Over Time}
We capture changes in attributes that always occur in a portrait (degrees of smiling) as well as in accessories or styles that are present in only some of the population at a given time.  

\subsubsection{Smiling in Portraiture}
A close observation of the decade average images in Figure~\ref{fig:decade_averages_teaser} reveals a change over time in the facial expression of portrait subjects.
In particular, today we take for granted that we are expected to smile when our picture is being taken; however, smiling at the camera was not always the norm.
In this section we attempt to quantify this change.

In her paper, the historian Kotchemidova studied the appearance of smiles in photographic portraits using the traditional historical methods of analyzing sample images manually~\cite{saycheese}. She reports that in the late 19th century people posing for photographs still followed the habits of painted portraiture subjects. These included keeping a serious expression since a smile was hard to maintain for as long as it took to paint a portrait.
Also, etiquette and beauty standards dictated that the mouth be kept small -- resulting in an instruction to ``say prunes" (rather than ``cheese") when photographed~\cite{saycheese}.
All of this changed during the 20th century when amateur photography became widespread.
In fact, Kotchemidova suggests that it was the attempt to associate photography with happy occasions like holidays and travel that led the photographic monopoly, Kodak, to educate the public through advertisements that the obvious expression one should assume in a snapshot is a smile. This multi-decade ad campaign was a great success.
By World War II, smiles were so widespread in portraiture that no one questioned whether photographs of the GIs sent to war should depict them with a smile~\cite{saycheese}.

\begin{figure}
\begin{center}
\includegraphics[width=0.9\linewidth]{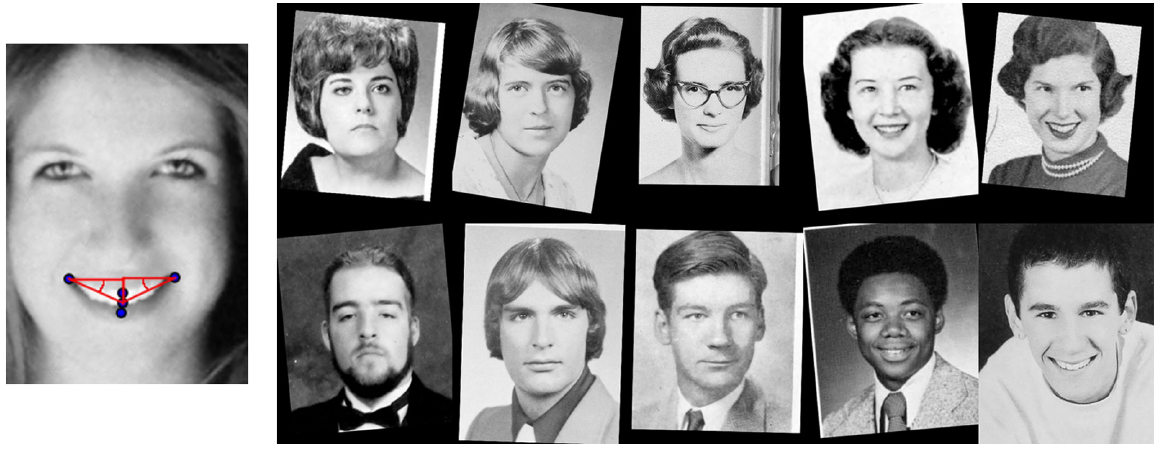}
\caption{Smile intensity metric. Left: the lip curvature metric is the average of the two marked angles. Right: women and men portraits sorted by increasing lip curvature.}
\label{fig:montage-curvature}
\includegraphics[width=0.8\linewidth]{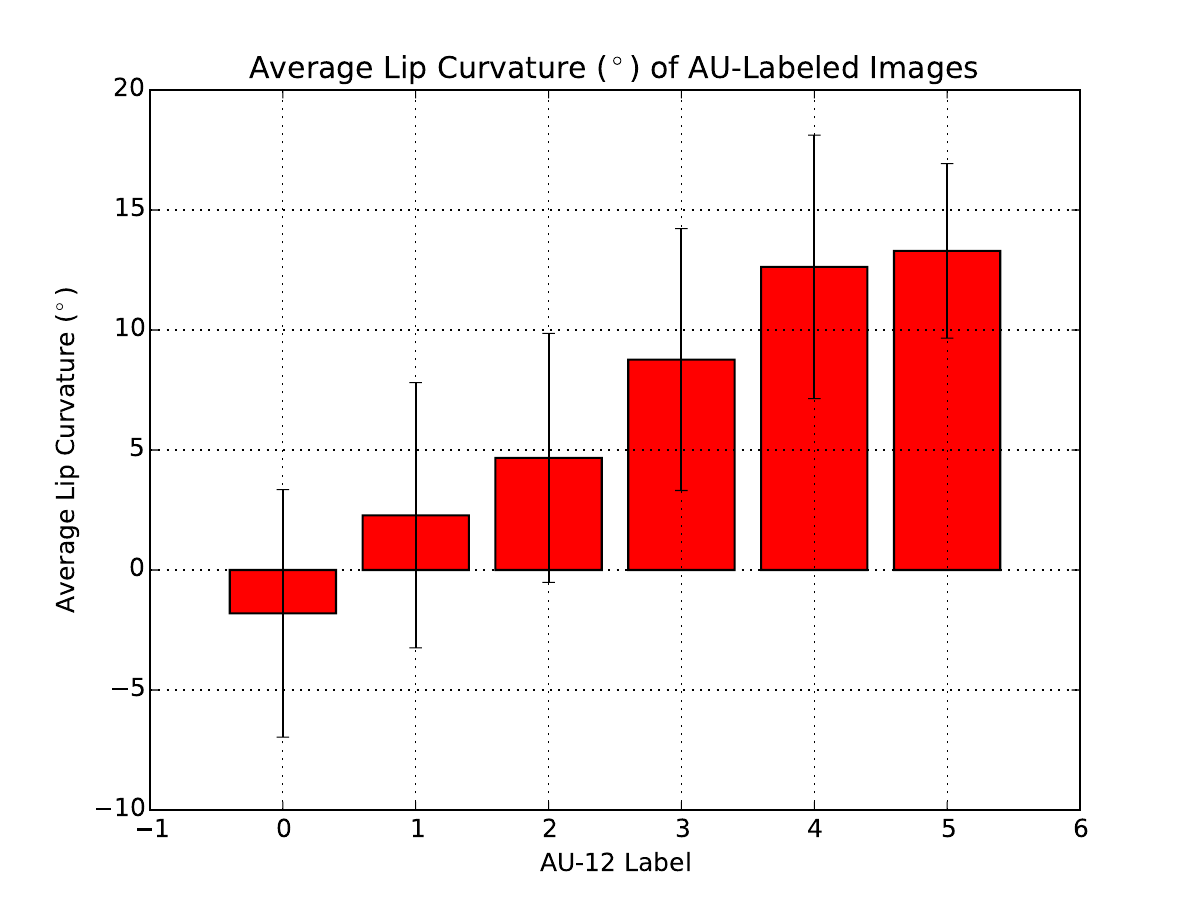}
\caption{Average lip curvature on BP4D data correlates with AU-12 labels which correspond to a contraction of the mouth muscles. Error bars denote standard deviation.}
\label{fig:BP4D-curvature}
\end{center}
\vspace{-0.2in}
\end{figure}

To verify the apparent trend in our average images and Kotchemidova's claims regarding the presence and extent of smiles in portrait photographs in a data-driven way, we devised a simple lip-curvature metric and applied it to our dataset.
We compute the lip curvature by taking the average of the two angles indicated in Figure~\ref{fig:montage-curvature} (Left) where the point that forms the hypotenuse of the triangle is the midpoint between the bottom of the top lip and the top of the bottom lip of the student.
The same facial keypoints were used here as in image alignment (see section~\ref{sec:data_prep}).
Figure~\ref{fig:montage-curvature} (Right) depicts a montage of students ordered in ascending order of lip curvature value from left to right.
Visually, the lip-curvature metric quantifies the smile intensities in our data in a meaningful way.

We verify that our metric generalizes beyond yearbook portraits by testing it on the BP4D-Spontaneous dataset that contains images of participants showing various degrees of facial expressions with ground truth labels of expression intensity~\cite{bp4d}.
BP4D uses labels drawn from the Facial Action Coding System, which is commonly used in facial expression analysis.
This system consists of Action Units (AU) that correspond to the intensity of contraction of various facial muscles.
Following previous work done on smile intensity estimation~\cite{girard2014automatic}, we compared our smile intensity metric with the activation of AU12 (Lip corner puller) as it corresponds to the contraction of muscles that raise the corners of the mouth into a smile.
A higher AU12 value represents a higher contraction of muscles around the corner of the mouth, resulting in a larger smile.
Figure~\ref{fig:BP4D-curvature} displays the average lip curvature for each value of AU12 for 3 male and 3 female subjects, corresponding to 2,500-3,000 samples for each AU12 value (0-5).
As the simple lip-curvature metric we used correlates with increasing AU12 values on BP4D images, it is a decent indicator for smile intensities beyond our Yearbook dataset.

\begin{figure}
\begin{center}
\vspace{-0.5in}
\includegraphics[width=0.9\linewidth]{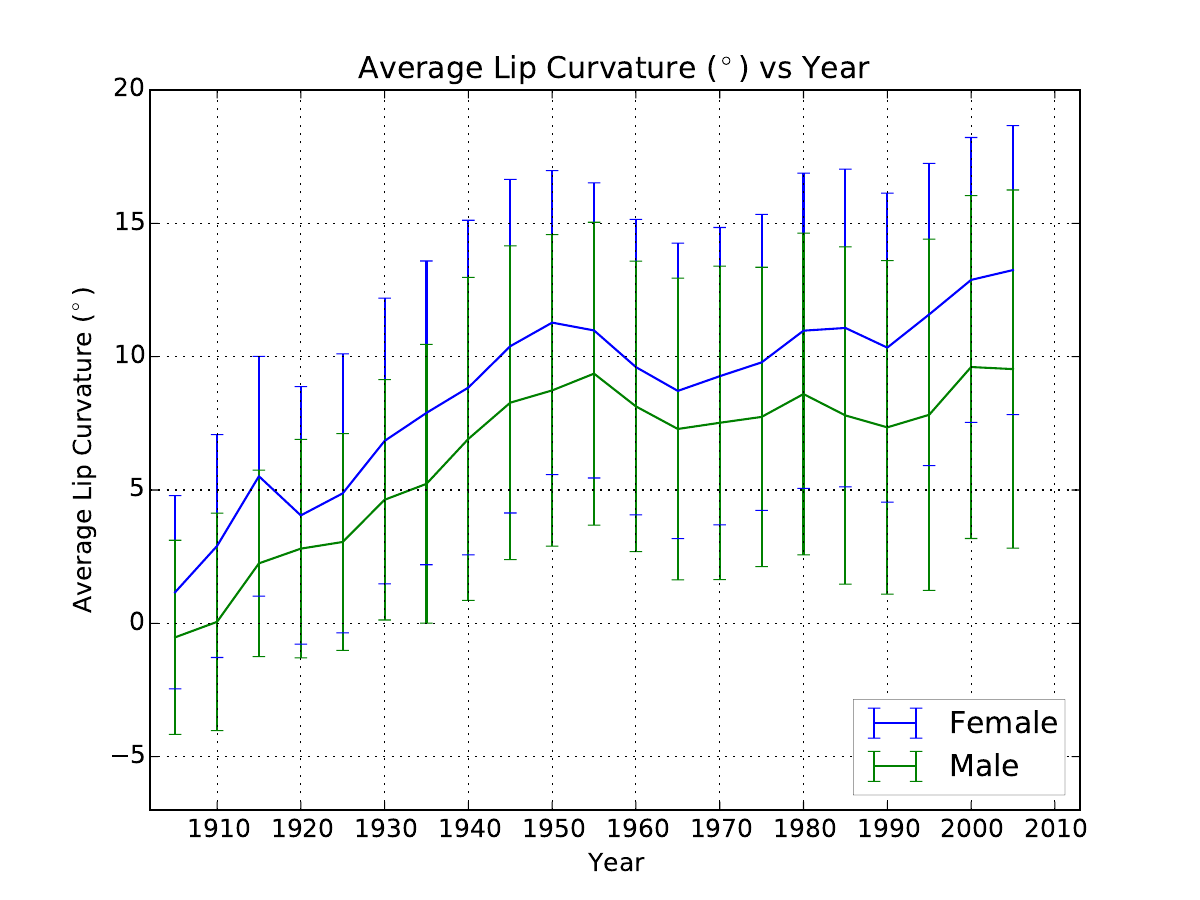}
 \caption{Smiles increasing over time, but on average, women smile more than men, across all decades: Male and female average lip curvature by year with one standard deviation error bars. Note the fall in smile extent from the 50s to the 60s, for which we did not find prior mention.}
 \label{fig:graph-years}
 \vspace{0.5cm}
\includegraphics[width=\linewidth]{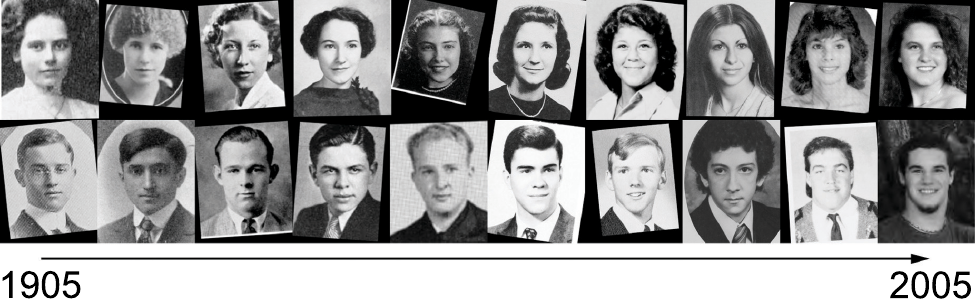}
 \caption{Images with the closest smile to the mean of that period (10-year bins from 1905 (left) to 2005 (right)). Note the increasing extent of smiles.}
 \label{fig:montage-years}
\end{center}
\vspace{-0.2in}
\end{figure}

Using our verified lip-curvature metric we plot the average smile intensities in our data over the past century in Figure~\ref{fig:graph-years}. Corresponding montages of smile intensities over the years are included in Figure~\ref{fig:montage-years}, where we picked the student with the smile intensity closest to the average for each 10-year bucket from 1905 to 2005.
These figures corroborate Kotchemidova's theory and demonstrate the rapid increase in the popularity and intensity of smiles in portraiture from the 1900s to the 1950s, a trend that still continues today; however, they also reveal another trend---women consistently smile more than men on average.
This phenomenon has been discussed extensively in the literature (see the review in~\cite{lafrance}), but until now required intensive manual annotation in order to discover and analyze.
For example, in her 1982 article Ragan manually analyzed 1,296 high school and university yearbooks and media files in order to reveal a similar result~\cite{Ragan}.
By use of a large historical dataset and a simple smile-detector we arrived at the same conclusion with a minimal amount of manual effort.

We note that smiles could also be detected using the expression recognition software from~\cite{intraface_final}.
However, this software was not publicly available at the time of our experiments.

\subsubsection{Glasses}
\begin{figure}
\begin{center}
\includegraphics[width=0.9\linewidth]{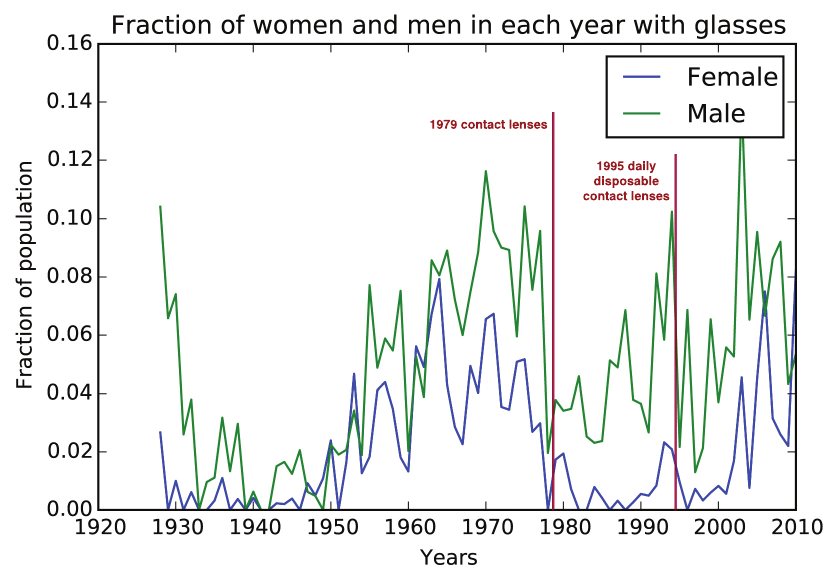}
 \caption{The use of glasses over time dips in correlation with advances in contact lenses, but glasses are consistently more popular among men.}
 \label{fig:glasses}
 \vspace{0.5cm}
 \includegraphics[width=0.9\linewidth]{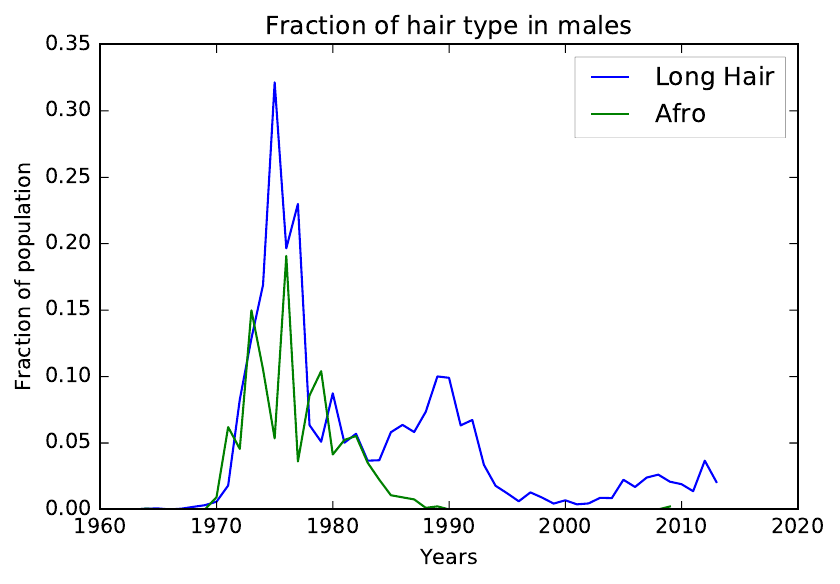}
 \caption{The fraction of male students with an ``afro" or long hair.}
 \label{fig:afros}
 \vspace{-0.2in}
 \end{center}
\end{figure}

Measuring the degree of smiles is easy to apply to each portrait in the collection since every subject exhibits \textit{some} degree of mouth curvature, albeit sometimes a negative one.
We now extend our study of trends to accessories and fashions that are only worn by a fraction of the population and that require a classification decision per portrait to determine if the specific style or accessory is exhibited.
We first study the usage of glasses by taking advantage of a small set of annotated celebrity portraits from the PubFig dataset~\cite{pubfig}.
We fine tune VGG~\cite{vgg}, a deep classification system pre-trained on ILSVRC~\cite{ILSVRC15}, on the celebrity portraits that are marked as wearing glasses.
We then apply the trained classifier to our Yearbook dataset to find persons wearing glasses in our data.
In Figure~\ref{fig:glasses} we graph the fraction of the student population that is wearing glasses for males and females over time.
It is interesting to note that glasses are more popular among male students, and to observe that the dips in glasses popularity correlate with the introduction of contact lenses.

\subsubsection{Men's Hairstyles post 1960}
The final trend we study is changes in men's hairstyles since the social movements of the 1960s which brought about long hair styles and ``afros".
Here we could not find an existing annotated dataset with appropriate annotations.
We therefore segmented out the hair in each portrait following~\cite{hairmask} and determined whether the depicted person had long hair or an afro by checking whether the segmentation map consists of hair under the depicted person's chin or high above his face, respectively.
(Note that this approach worked well on our data due to the lack of facial hair among most high school students).
Unfortunately, due to the low resolution of some of the portraits in our dataset the fully-automatic approach was not accurate enough and extra manual filtering was required.
Figure~\ref{fig:afros} shows the fraction of the population with these hairstyles after a manual process of removing false positives and adding some false negatives to our classifications.
We note that our findings corroborate other sources~\cite{afro,hairencyclopedia} which claim that the afro hairstyle was predominantly popular from the late 1960s through the late 1970s after which many individuals switched to a more styled version of the natural hairdo.

\vspace{-0.1in}
\subsection{Mining for Date-Specific Patterns}
\label{sec:date-specific}
The average images of each decade from Figure~\ref{fig:decade_averages_teaser} show us the main modes of the styles of each decade.
However, in each time period or even classroom not every one shares the same style.
In fact, we would expect to find several representative and visually discriminative features for every decade.
These are the things that make us immediately recognize a particular style as ``20s" or ``60s", for example, and allow humans to effortlessly guess the decade in which a portrait was taken.
They are also the things that are usually hard to put into writing and require a visual aid when describing; this makes them excellent candidates for data-driven methods.

We find the most representative women's styles in hair and facial accessories for each decade using a discriminative mode seeking algorithm~\cite{doersch2013mid} on yearbook portraits cropped to contain only the face and hair.
Since our portraits are aligned, we can treat them as a whole rather than look for mid-level representative patches as has been done in previous work~\cite{doersch2012what,doersch2013mid}.
The output of the discriminative mode seeking algorithm is a set of detectors and their detected portraits that make up the visual clusters for each decade.
We sort these clusters according to how discriminative they are, specifically, how many portraits they contain in the top 20 detections from the target decade versus other decades.
In order to ensure a good visual coverage of the target decade, we remove clusters that include in their top 60 detections more than 6 portraits (10\%) that were already represented by a higher ranking cluster.

\begin{figure*}
\begin{center}
\includegraphics[width=\linewidth]{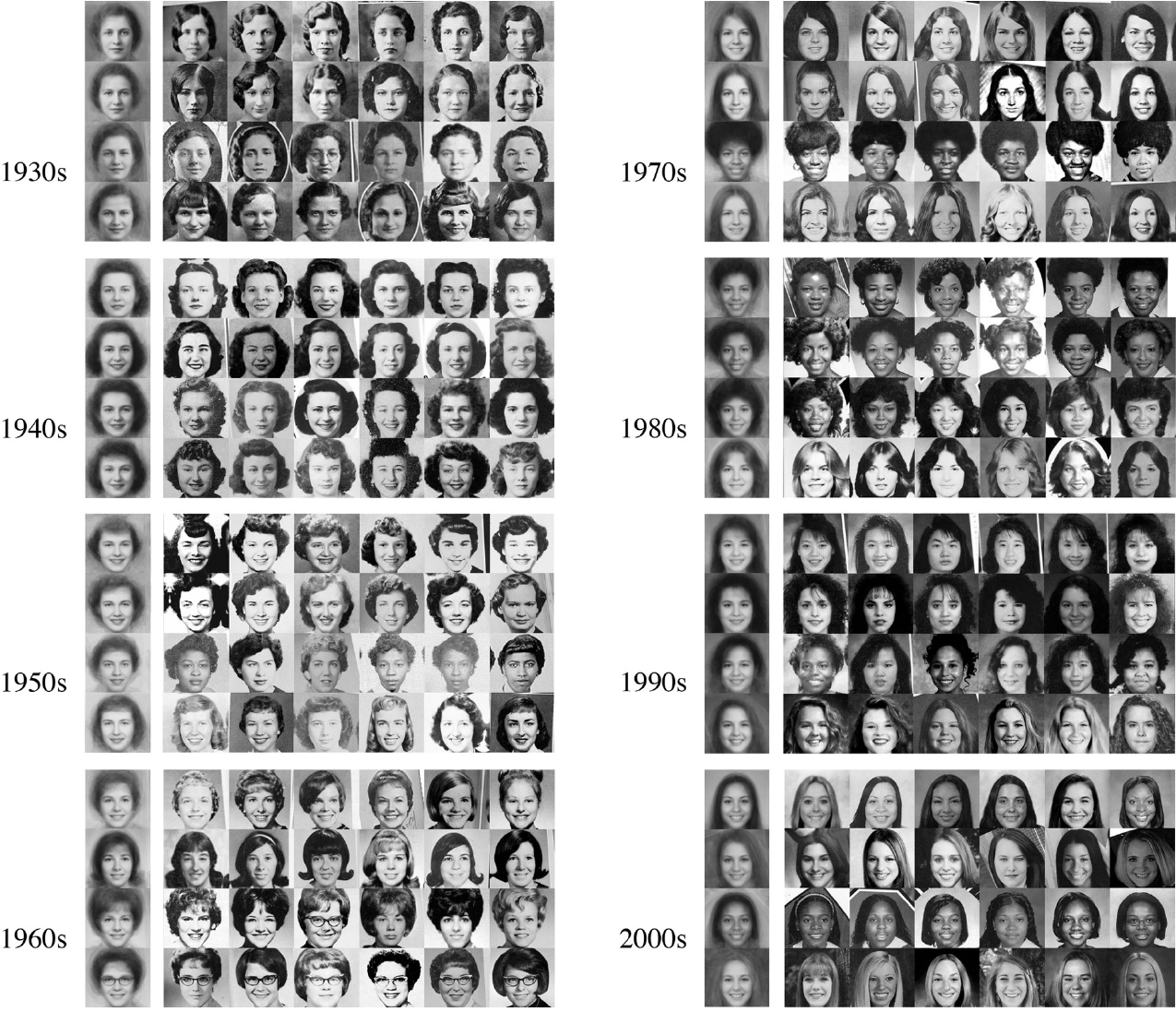}
\end{center}
   \caption{Discriminative clusters of high school girls' styles from each decade of the 20th century. Each row corresponds to a single detector and the cluster of its top 6 detections over the entire dataset. Only one girl per graduating class is shown in the top detections. The left-most entry in each row displays the cluster average. Note that the clusters correspond to the quintessential hair and accessory styles of each decade. Notable examples according to the Encyclopedia of Hair~\cite{hairencyclopedia} are: The finger waves of the 30s. The pin curls of the 40s and 50s. The bob, ``winged" flip, bubble cut and signature glasses of the 60s. The long hair, afros and bouffants of the 70s. The perms and bangs of the 80s and 90s and the straight long hair fashionable in the 2000s. These decade-specific fashions emerge from the data in a weakly-supervised, data-driven process.}
\label{fig:detections_full_60s_exclusions}
\end{figure*}

Figure~\ref{fig:detections_full_60s_exclusions} displays the four most representative women's hair and eyeglass styles of each decade from the 1930s until the 2000s.
Each row corresponds to a visual cluster in that decade.
The left-most entry in the row is the cluster average, and to its right we display the top 6 portrait detections of the discriminative detector that created the cluster.
We only display a single woman from each graduating class in order to ensure that the affinity within each cluster is not due to biases in the data that result from the photographic or scanning artifacts of each physical yearbook.
Looking at Figure~\ref{fig:detections_full_60s_exclusions}, we get an immediate sense of the attributes that make each decade's style distinctive.
For example, the particular style of curly bangs of the 40s or the ``winged" flip hairstyle of the 60s~\cite{hairencyclopedia}.
Finding and categorizing these manually would be painstaking work.
With our large dataset these attributes emerge from the data by using only the year-label supervision.

\section{Dating Historical Images}
\label{sec:dating}

In Section~\ref{sec:date-specific} we found distinctive visual patterns that occur in different decades.
Here we ask whether there are enough decade-specific visual patterns to be able to predict the year in which a portrait of a face was taken.
We refer to this task as the \textit{portrait dating} problem.

We extend the work of Palermo et al.~\cite{dating} in dating color photographs to the realm of black and white portraiture photography where we cannot rely on the changes in image color profiles over time.
We choose to train a deep neural network model for dating portraits based on the recent success of such models for other visual recognition tasks~\cite{vgg}.
While the portrait dating problem can be cast into a regression framework, a standard regression formulation models the data with a Gaussian distribution, eliminating the possibility of multiple modes.
We therefore choose to model the problem as classification.

We pose the task of dating the portraits of female and male students as an 83-way year-classification task between the years 1928 and 2010, the years for which we have more than 30 female and male images per year. 
Separate classifiers are trained for each gender to discourage the model from using low-level image artifacts as a discriminatory signal.
The models trained on women and men are referred to as the \textit{women's model} and \textit{men's model} respectively.
We evaluate our model on a subset of images drawn from the Yearbook dataset, the \textit{Yearbook test set}, which is also divided by gender.
To assess the generalization capability of our dating model, we conduct experiments including testing the model on yearbook photos of the opposite gender, evaluating the model on a small set of celebrity photos, and training a classifier on random background crops. 

\subsubsection{Dating Yearbook Portraits}
Our date-prediction model is based on the VGG-16 model~\cite{vgg} that was pre-trained on the ILSVRC benchmark image classification task~\cite{ILSVRC15}. 
The network implementation and training procedure are detailed at the end of this section.
In Table~\ref{tbl:classification}, we present results for two network models and a baseline:
\begin{itemize}
\item \textbf{Partial FT}: freeze the weights of all convolutional layers and train only the fully connected layers and final classification layer of the network.
\item \textbf{Full FT}: fine-tune all layers of the network.
\item \textbf{Chance}: a baseline defined as the inverse of the number of classes.
\end{itemize}

Results for the Yearbook test set for each gender are shown in column \emph{Test}.
Fine-tuning the full network on the Yearbook data provides a performance boost over partial fine-tuning, indicating that the convolutional filters in the lower layers can be effectively tuned to Yearbook-specific features.
Quantitatively, 65.3\% of the women and 46.4\% of the men test images are classified within 5 years of the true year.
To investigate the large gap in performance between the men's and women's models, we trained models for both genders on the easier problem of 10-way ``decade" classification.
This classifier achieves 61.0\% accuracy when trained on the women's data, but only 44.3\% when trained on the men's data.
We conclude that there is simply less discriminative signal present in the images of men, and hypothesize that men's appearances change less over time, resulting in few time-specific semantic features. 
For example, the average images in Figure~\ref{fig:decade_averages_teaser} demonstrate that sporting short hair and a suit was the default fashion choice across all decades.

\begin{figure}
\begin{center}
\includegraphics[width=0.9\linewidth]{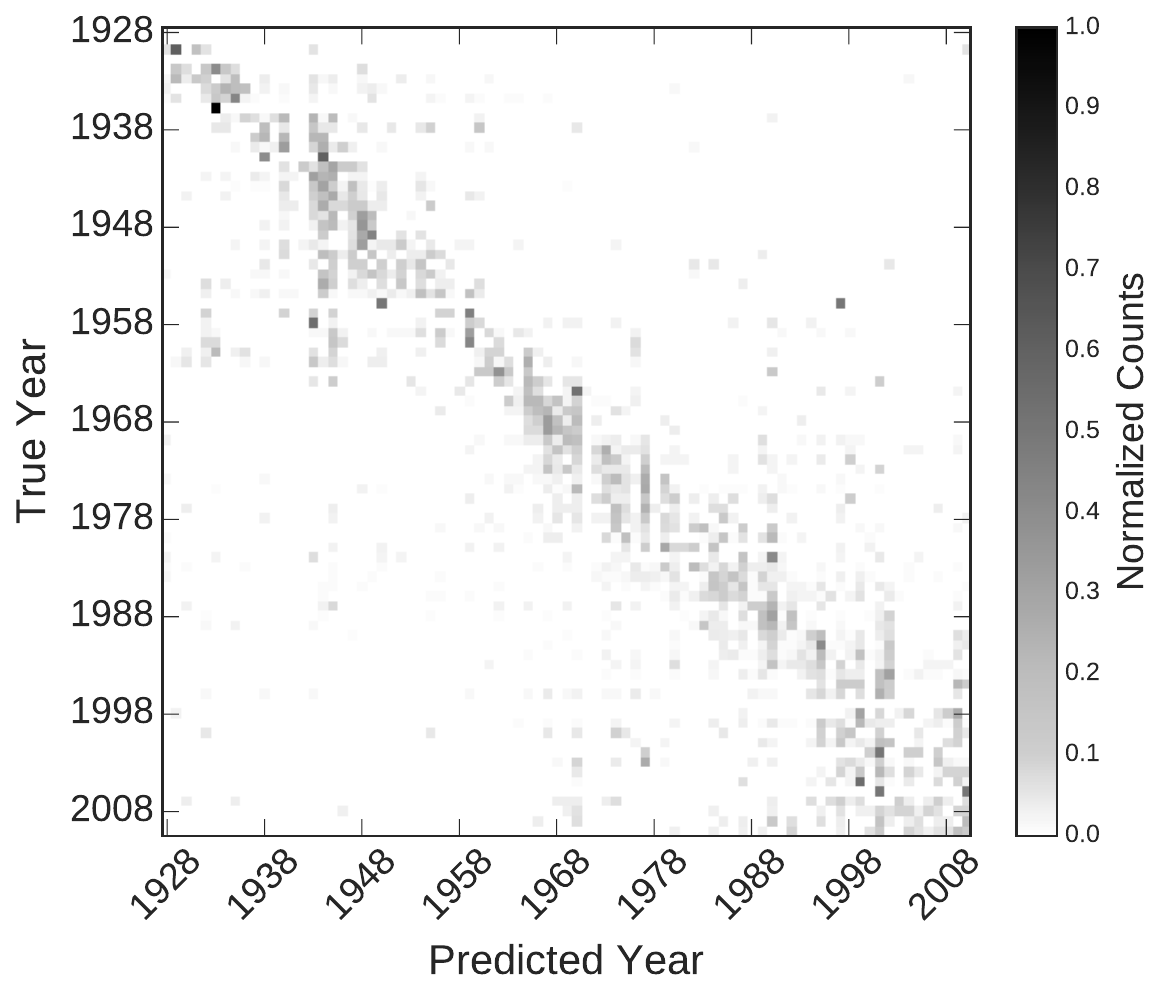}
\end{center}
   \caption{Confusion matrix for the fully fine-tuned women's model evaluated on the Yearbook women test set, with each row normalized by the number of images in that year. Darker off-diagonal regions indicate more confusion. The mostly diagonal structure demonstrates that confusion mostly occurs between neighboring years, indicating that the dating model can distinguish between time periods.} 
\label{fig:confmat}
\vspace{-0.5cm}
\end{figure}

For the women's model, full fine-tuning improves the L1 median error in addition to the accuracy on the women's test set.
Furthermore, the confusion matrix visualized in Figure~\ref{fig:confmat} reveals that the predictions are rarely far off the mark.
The diagonal structure indicates that most of the confusion occurs between neighboring years, matching our intuition that visual trends such as hairstyle transcend the single-year boundary.
\begin{table}
\small
\setlength\tabcolsep{3.8pt}
\centering
\begin{tabular}{clccccccc}
\toprule

& & \multicolumn{3}{c}{Accuracy [\%]} & &\multicolumn{3}{c}{L1 Med Error [yrs]} \\
\cmidrule[0.5pt]{3-5}
\cmidrule[0.5pt]{7-9}
& Model & Test & Other & Celeb && Test & Other & Celeb \\
\midrule
\multirow{3}{*}{\rotatebox[origin=c]{90}{Women}}
& Chance & 1.2 & 1.2 & 1.2 && - & - & -  \\
& Partial FT & 8.1  & 3.4 & 0  && 5  & 11   & 27  \\
& Full FT  & 10.9  & 4.0 & 5.2 && 4  & 8 & 17  \\
\midrule
\multirow{3}{*}{\rotatebox[origin=c]{90}{Men} } 
& Chance & 1.2  & 1.2 & 1.2 && - & - & - \\
& Partial FT & 4.8 & 3.2 & 0 && 6 &10   & 20  \\
& Full FT & 5.5   & 3.7 & 0 && 6  & 10 & 20  \\
\bottomrule
\vspace{1pt}
\end{tabular}
\caption{Classification accuracy and L1 median error for the Yearbook men's and women's classification models on the task of 83-way year classification between years 1928-2010. ``Test" refers to the test set of the same gender, ``Other" refers to the test set of the opposite gender, ``Celeb" refers to the celebrity test set.} 
\label{tbl:classification}
\vspace{-0.4cm}
\end{table}

\subsubsection{Generalization}

\begin{figure*}
\centering
\begin{tabular}{c c c}
\includegraphics[width=0.32\linewidth]{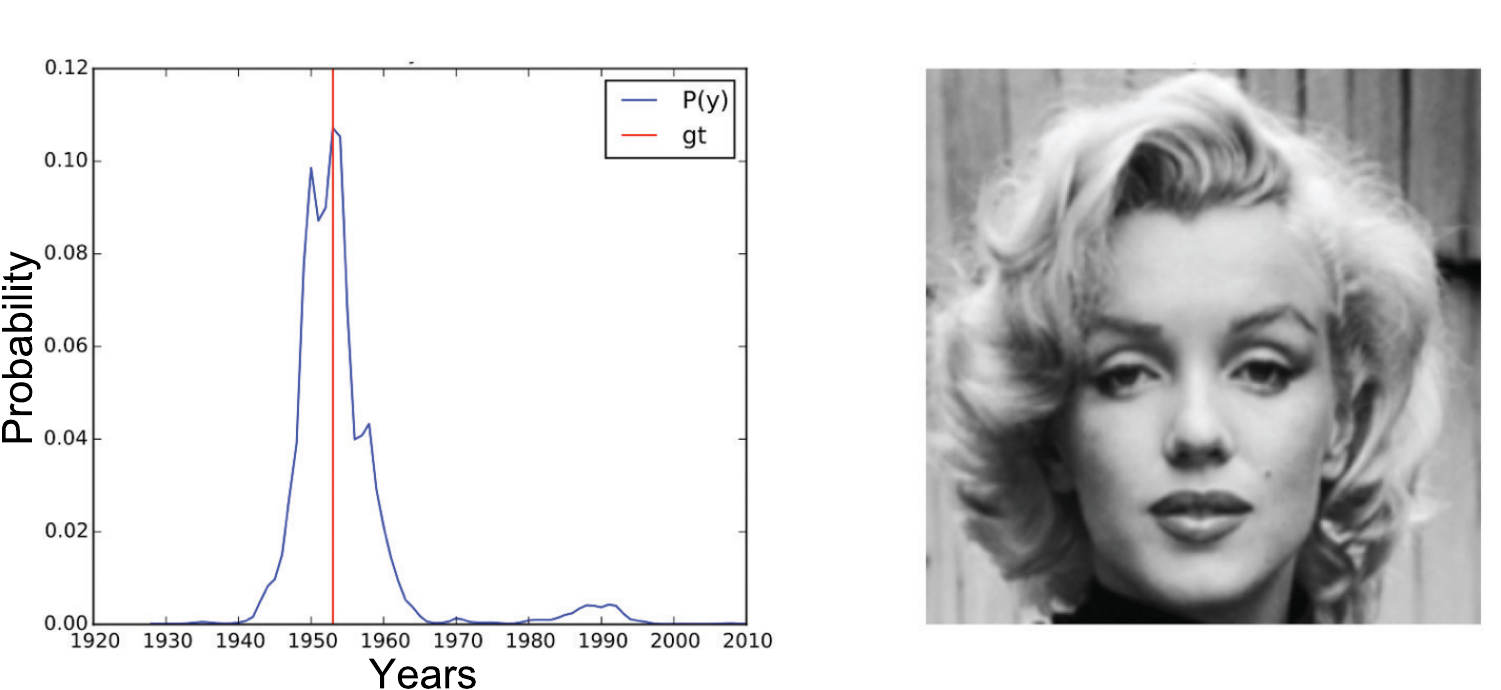}
& \includegraphics[width=0.32\linewidth]{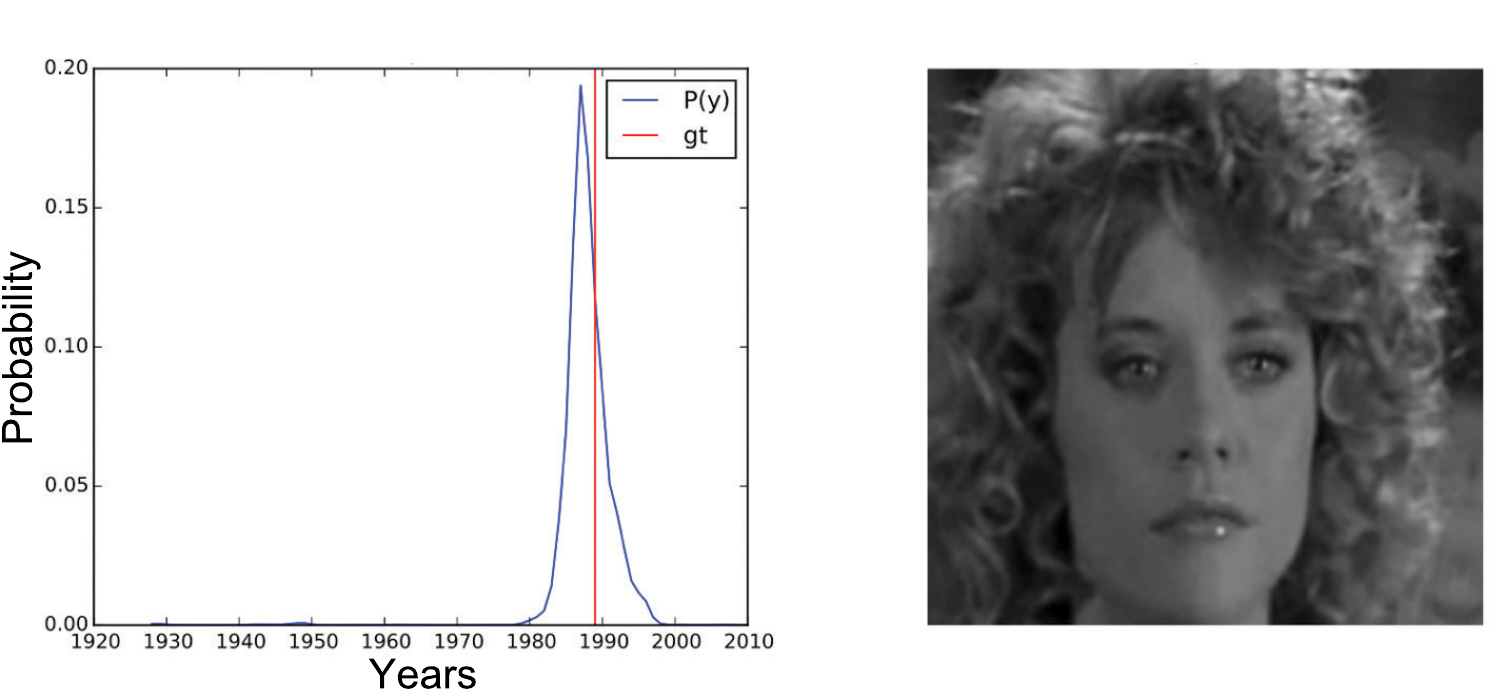}
& \includegraphics[width=0.32\linewidth]{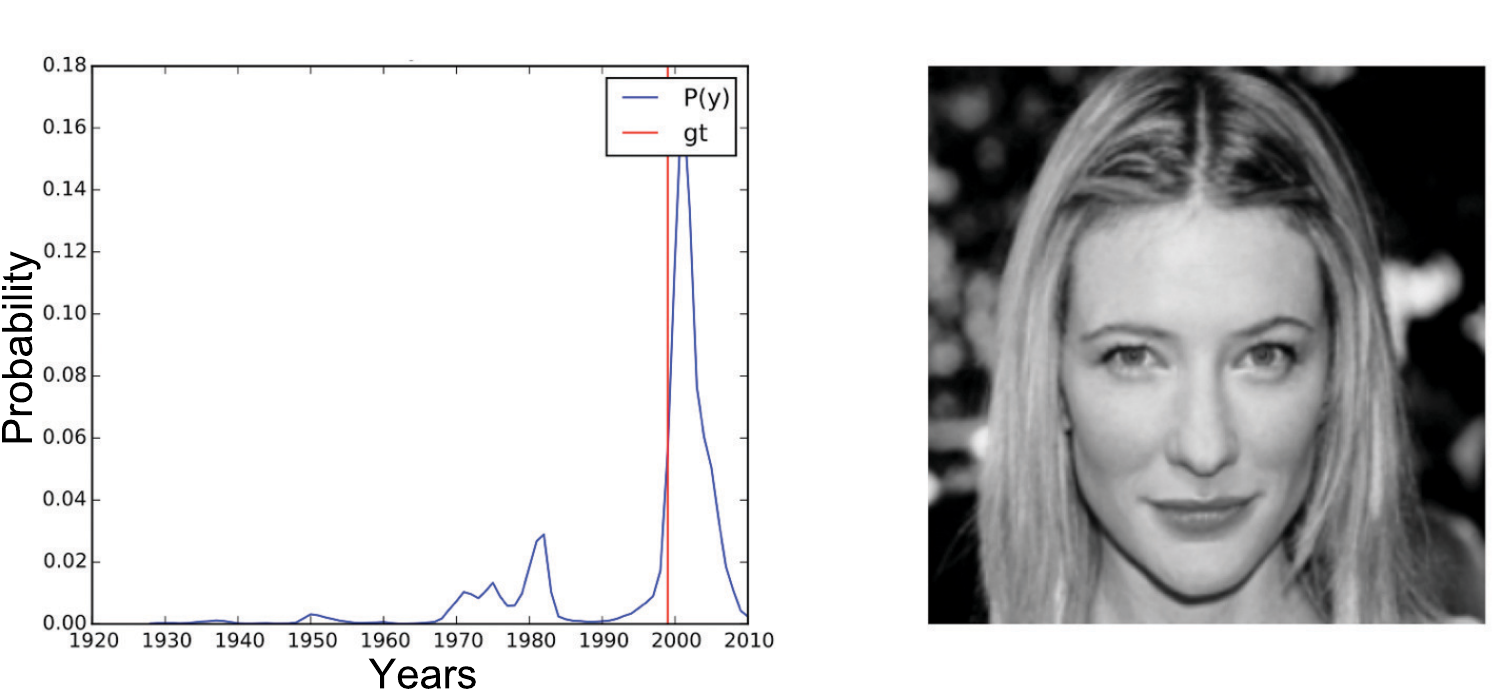}
\\ {\scriptsize(a) Ground truth year: $1953$. Predicted $1953$.} & {\scriptsize(b) Ground truth year: $1989$. Predicted $1987$.} & {\scriptsize(c) Ground truth year: $1999$. Predicted $2001$.}  \\
\end {tabular}
   \caption{The dating model generalizes somewhat to celebrity glamour shots, a significant domain shift from the yearbook photos on which the model was trained. Shown are good celebrity dating predictions. Red indicates the ground truth year. Blue indicates the prediction distribution.}
\label{fig:celeb-pred}
\vspace{-0.1cm}
\end{figure*}

The success in dating yearbook portraits may be misleading since there are biases in the Yearbook dataset that the network can exploit, such as similar backgrounds and low-level image statistics.
To determine the potential usefulness of such low-level cues, we train a classification model on 32 by 32 pixel crops of portrait background (crops are taken from the corners of each image in the Yearbook women training set). This model achieves 2.8\% accuracy, and 24.1\% accuracy within five years. Such poor performance demonstrates that low-level image statistics and portrait backgrounds are not sufficient to date the portraits.

To further test the generalization power of our portrait dating model, we test it on two different datasets not seen during training. First, we test each model on the Yearbook photos of the opposite gender than those with which the model was trained. High performance across genders would indicate that the model leverages low-level statistics common across all the Yearbook photos. Second, we test each model on the \textit{celebrity test set} -- a small set of 100 gray-scale head shots of celebrities (58 female, 42 male), annotated with year labels, that we cropped and aligned to the Yearbook images. High performance across students and celebrities would indicate that the model uses higher-level cues such as hairstyle to perform the dating task. 
The results for these two generalization experiments are presented in Table~\ref{tbl:classification}, in columns \emph{Other} and \emph{Celeb} respectively.

For both the men's and women's models, performance on the yearbook photos of opposite gender is substantially worse than for the gender on which the model was trained, thus low-level image statistics cannot account for the success of the dating model. 
The fully fine-tuned women's model greatly improves performance on the celebrity test set compared to the baselines, suggesting that generalizable features are learned from the Yearbook data. 

The performance gap between the Yearbook test photos and the celebrities for both models indicates that some cues used by the model are Yearbook-specific. 
This reduced performance may be due to the domain shift between portraits of high school students and celebrity glamour shots; celebrity hairstyles can be quite different than those of the general public.
Additionally, our celebrity test set may simply be too small to serve as an informative test set.
However, while dating does not generalize well for all celebrities, approximately 40\% of the L1 errors on female celebrities are less than a decade and most predictions are within two decades of the ground truth year.
Figure~\ref{fig:celeb-pred} displays individual good predictions.

\subsubsection{Implementation Details} For the dating task, we use portraits that were cropped to the face and hair alone.
The yearbook test set consists of approximately 30\% of the portraits taken between 1982 and 2010: 4,227 women and 4,489 men. The remaining 80\% of images are used for training and validation: 15,370 women and 13,184 men.
To minimize training biases due to photographic and scanning artifacts, we separate test and training images drawn from the same school by at least a decade.
To further minimize these biases, we use the built-in Photoshop noise reduction filter on all the Yearbook images and resize them to 96 by 96 pixels.
In all of our experiments, we use the Caffe~\cite{jia2014caffe} framework for training deep learning models.
For the classification model, we use the VGG network architecture~\cite{vgg} that was pre-trained on the ILSVRC benchmark image classification task~\cite{ILSVRC15}.
We modify the fully connected layers to accommodate 96$px$ inputs, and add an 83-output classification layer followed by a softmax cross-entropy loss.
All networks are trained for $5K$ iterations of mini-batch size 64 with horizontal mirroring data augmentation, using SGD with learning rate $0.001$, momentum $0.9$, and weight decay $5e$-$4$.\footnote{Code to reproduce our results is available at \url{https://github.com/katerakelly/yearbook-dating}.}

\section{What time specific patterns is the classifier using for dating?}
In section~\ref{sec:dating} we demonstrated that it is possible to train a classifier to guess the date in which a portrait was taken.
But what is the classifier doing? What time-specific visual features is it picking up on?
In this section we visualize which pixels are responsible for a given dating decision.
The latent representations at the intermediate layers of a feed-forward convolutional neural network $f$ are grouped into spatial locations, such that several features are activated at each spatial location in different feature channels.
While the ensemble of hidden activations learns a large, distributed code for the training data, it is never used in its entirety to represent a single input -- different inputs take different paths through the network during inference. 
Therefore, for a single input we can safely disable the spatial locations throughout the network that are not part of the path for this specific input while keeping the same output.
This process of removing unused locations that do not participate in the computation of a particular $y=f(x)$ allows us to visualize the parts of the input image that do.
Next we present an algorithm that implements this process.

\subsubsection{Top-Down Selection of Spatial Units}
\label{sec:technical}
Our goal is to ask `What parts of the image were used to make \textit{this} decision?'.
We therefore would like to maintain the same output distribution while removing unnecessary units.
Given an input $x$ we compute its resulting probabilistic output $y=f(x)$ by running a forward pass over the network.
Here the output $y$ is a vector with $n$ entries corresponding to $n$ years, where each entry contains the probability that a given photograph is from a given year.
We then run a single top-down optimization pass where we disable units in spatial locations that are not needed to produce the probability distribution $y$.
Since our goal is to maintain the same output distribution, we use the KL divergence, a distance measure between two distributions, as our objective function.
Specifically, we define the objective to be the KL divergence $D_{KL}(y||\hat y_l)$ of the predicted output $\hat{y_l}$ after spatial unit removal at layer $l$ from the true final output distribution of the network $y$:
\begin{equation}
D_{KL}(y || \hat{y_l}) = \sum_c( y_c\log\frac{y_{c}}{\hat{y}_{lc}} ),
\label{eq:KLD}
\end{equation}
where $c$ refers to a single entry in the probabilistic output of the CNN (or a single class).

We minimize the KL divergence via the following optimization that forces the network to keep only a sparse set of active units, while maintaining the same output distribution:

\begin{equation}
\begin{aligned}
& \underset{M_l \in \{0,1\}^N}{\text{minimize}}
& & D_{KL}(y || \hat{y_l}) \\
& \text{subject to}
& & \|M_l\|_0 \leq s_l N.
\end{aligned}
\end{equation}
Where $M_l$ is a $2D$ binary mask that disables spatial units at the input to layer $l$ where its elements are $0$, and $s_l$ is the desired sparsity percentage over the $N$ spatial units in layer $l$.
For simplicity, we use the same fixed sparsity percentage throughout all layers.

To perform the above optimization we use a greedy algorithm that traverses the network once from top to bottom and minimizes the objective with respect to the constraint at every layer.
For each layer, we iterate over all spatial locations of its input feature map and output a binary mask $M_l$ which removes all spatial units that are not necessary for computing the output distribution $y$.
We jointly disable all features grouped at a single spatial location (all channels for a single location).
Note that when $M$ does not remove any spatial locations this objective is minimized but the sparsity constraint is violated.
We therefore start from a full mask $M$ of all 1's for each layer and remove (zero out) those locations whose removal increases the value of the objective as little as possible.
This approach is similar to Orthogonal Matching Pursuit~\cite{Pati93orthogonalmatching}, although there the objective is usually a Euclidean distance.
For a detailed description, refer to Algorithm~\ref{algo:1}.

\begin{algorithm}
\small
\caption{Greedy top-down selection of spatial units}
\begin{algorithmic}[1]
\FOR{each layer $l$}
\STATE{Start from a mask $M_l$ of all 1's}
\WHILE{number of active spatial units $>s_lN$}
\FOR{each spatial location $i$}
\STATE{Zero out $i$ in layer $l$}
\STATE{Run a forward pass from $l$, zeroing locations in higher layers $h$ that were previously disabled}
\STATE{Compute predicted output $\hat{y_l}$ and loss $D_{KL}(y || \hat{y_l})$}
\ENDFOR
\STATE{Zero out the spatial location with the smallest increase in the loss function: $D_{KL}(y || \hat{y_l})$}
\ENDWHILE
\ENDFOR
\end{algorithmic}
\label{algo:1}
\end{algorithm}

\subsubsection{Gradient Approximation}
The iterative greedy algorithm of removing one spatial location at a time at each layer is too slow to run in practice for lower-level layers of the CNN since it iterates over all spatial locations of the feature map for every spatial unit it disables.
To make the optimization faster we first present an alternative interpretation of Algorithm~\ref{algo:1} and then show how to approximate the expected change in loss for any unit using a single backward pass through the network.

At each step of Algorithm \ref{algo:1} we find a spatial single unit $i$, which when set to $0$ increases the loss the least.
This increase in loss can be measured as follows:
\begin{equation}
  d_i = D_{KL}(y || \hat{y_l}^\prime) - D_{KL}(y || \hat{y_l}), \label{eq:diff}
\end{equation}
where $\hat{y_l}$ and $\hat{y_l}^\prime$ are at a single spatial location $i$ that is zeroed out in $\hat{y_l}^\prime$.
Equation \ref{eq:diff} can be thought of as a finite-difference approximation with ${d_i = z_{l,i} \frac{\partial}{\partial z_{l,i}} D_{KL}(y || \hat y_l)}$ (though the difference here may be large), and can thus be approximated by the product of the gradient of the KL divergence objective function ${\frac{\partial}{\partial z_{l,i}} D_{KL}(y || \hat y_l)}$ and the value of the input activations $z_{l,i}$ of layer $l$.
While this linear approximation is crude it works well in practice and only requires a single backward pass through the network, replacing lines $4-8$ in Algorithm~\ref{algo:1}.

Note that when the two distributions, $y$ and $\hat{y_l}$, are equal the gradient of the objective is zero.
In implementing this approximation we therefore reverse the direction of the optimization -- we start from a mask $M_l$ of all $0$'s (in line $2$ of Algorithm~\ref{algo:1}) and add the subset of spatial units that are necessary to maintain the output distribution.

\subsubsection{Experimental Setup}
We run our spatial unit selection algorithm on the dating classification network that we fine tuned from the ILSVRC-trained VGG~\cite{vgg} as in section~\ref{sec:dating}.
The VGG network consists of a deep stack of convolutional layers and two fully-connected layers at the top.
While the algorithm runs out-of-the-box on VGG, the fully connected layers discard the spatial component of their input feature maps that was maintained throughout the convolutional stack.
We therefore modify the network where, following Long et al.~\cite{long_shelhamer_fcn}, we replace the fully-connected layers with convolutional ones creating a fully convolutional version of VGG.
Unlike~\cite{long_shelhamer_fcn}, we use $1\times1$ convolutions to replace all upper layers, reducing the parameters of the model as well as the receptive field size of each unit.
This allows us to treat each image-pixel as an independent predictor for image-class $c$.
Since we do not have pixel-level ground truth annotations for the image-level dating task, we take the final image-level date prediction to be the average over all spatial predictions. In our experiments we use a fixed sparsity $s_l = 20\%$ for all layers.

\subsubsection{Quantitative Evaluation}

\label{sec:eval}
\begin{figure}
\centering
\begin{tabular}{c c}
        \includegraphics[width=0.25\linewidth]{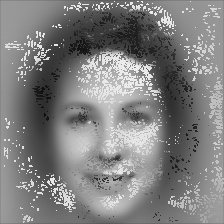}
        & \includegraphics[width=0.25\linewidth]{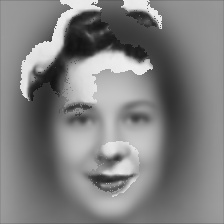}
        \\
        {\scriptsize Simonyan et al.~\cite{Simonyan14a}} & {\scriptsize Ours}
\end{tabular}
 \caption{Discriminative regions for a $1940$ portrait found by each method, overlaid on the mean training image.}
\label{fig:data}
\end{figure}

Unfortunately, network visualization papers have historically only provided qualitative evaluations of their results.
A noteworthy exception is~\cite{SamekBMBM15} who propose a method based on region perturbation for evaluating pixel relevance heatmaps.
We provide a simpler quantitative measure of the discriminativeness of the discovered regions by testing how a pre-trained network could predict the year label of Yearbook images \textbf{only} from the discovered elements.
To this end, we use a network that has been fine-tuned on the original training data to classify the pixel-level discriminative regions for different methods.
For each test instance, we start with the training-set mean image and add the color values of the discovered regions (see Figure~\ref{fig:data}).
Table~\ref{table:quant} shows the accuracy of our approach compared with Simonyan et al.~\cite{Simonyan14a} on the resulting images.
As expected, our method achieves a higher classification accuracy since it retains more discriminative elements.

\begin{table}
\small
\begin{center}
\begin{tabular}{|l|c|c|c|}
\hline
Method & Accuracy & Avg L1 Error & Med L1 Error\\
\hline\hline
\cite{Simonyan14a} & 0.017 & 24.0 &  20.0 \\
ours & \textbf{0.033} & \textbf{18.1} & \textbf{11.0} \\
\hline
\end{tabular}
\end{center}
\caption{\mbox{Classification accuracy and errors on visual elements.}}
\label{table:quant}
\vspace{-0.2in}
\end{table}

\subsubsection{Qualitative Evaluation}
\label{sec:necessary}

\begin{figure*}
\centering
        \includegraphics[width=\textwidth]{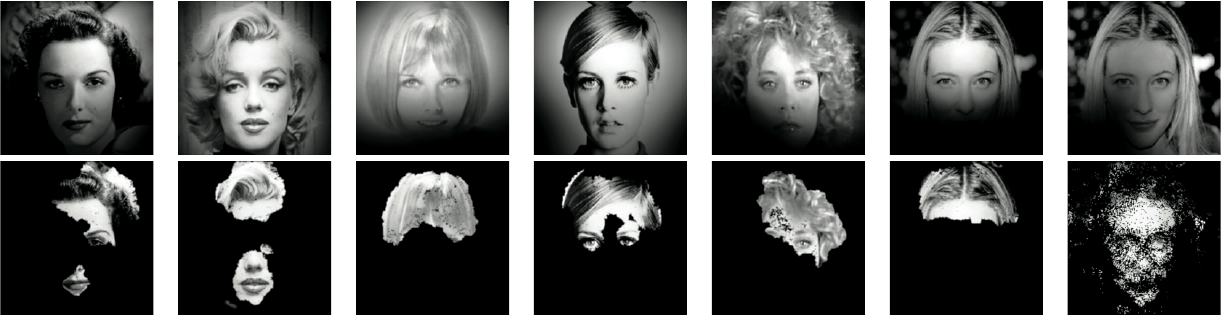} 
        \\
        {\scriptsize(a) $1945$} \hspace{40pt} {\scriptsize(b) $1953$}  \hspace{40pt} {\scriptsize(c) $1966$}  \hspace{50pt}{\scriptsize(d) $1966$} \ \hspace{40pt} {\scriptsize (e) $1989$}  \hspace{40pt} {\scriptsize(f)  $1999$} \hspace{40pt} {\scriptsize(g) $1999$,~\cite{Simonyan14a}}  \\
\caption{(a)-(f) Results on celebrity portraits from different eras. (g) In comparison with column (f),~\cite{Simonyan14a} tends to focus on the middle of the object, the nose and forehead. Rows represent the selected spatial units in the inputs to layers $pool_5$ (top) and $conv_1$ (bottom). While the unit selection process is a hard-selection, we shade the receptive field of each unit in the $pool_5$ layer using a tent filter for displaying purposes.}
\label{fig:backpropagate}
\end{figure*}

The results of applying our spatial-unit selection algorithm are shown in Figure~\ref{fig:backpropagate} and compared to the results of the Simonyan et al.~\cite{Simonyan14a} method.
Our algorithm extracts image parts that are meaningful for dating such as $40$'s and $50$'s dark lipstick, $60$'s flat bangs, $80$'s curls and $90$'s hair partings.
Referring back to Figure~\ref{fig:detections_full_60s_exclusions}, we have verified that we can localize the visual elements that resulted in these full image decade clusters.
In comparison, the Simonyan et al. method tends to pick out the center of the object, here the forehead and nose of the depicted person, which is less relevant for predicting the era of the photograph.
The images used here are all correctly predicted images from the \textit{unseen} set of female celebrity portraits.

\section{Conclusion}
In this paper, we presented a large-scale historical image dataset of yearbook portraits, which we have made publicly available.
These provide us with a unique opportunity to observe how fashions and habits change over time in a restricted, fixed visual framework.
We demonstrated the use of various techniques for mining visual patterns and trends in the data that significantly decrease the time and effort needed to arrive at the type of conclusions often researched in the humanities.
We showed how deep learning techniques can leverage the time-specific visual information in a single facial image to date portraits with great accuracy.
Moreover, we presented a technique to visualize which parts of the image are used in dating the portraits thus finding the discriminative visual elements of each time period.

Through the process of working with historical images we often pushed the current state-of-the-art computer vision techniques to their limits.
While some automatic methods, such as face detection, are robust enough for low resolution and low quality scans, there is much room for the improvement of other methods that are often only tested on high quality imagery.
Some examples include automatic figure-ground and hair segmentation methods, facial keypoint detection that captures the full facial mask, 3D alignment of faces that respects hair and accessories, accurate pose estimation and the detection of face attributes and accessories such as long hair and jewelry.
Finally, our main challenge working with CNNs was ensuring that they do not memorize semantically unimportant artifacts such as portrait backgrounds and noise.

Much remains to be done in the application of machine learning techniques to visual historical datasets, and in particular the one at hand.
For example, historical yearbook portraits can be used to characterize the spread of styles over spatio-temporal domains and the influence of celebrity styles on the public, to discover the cycle-length of fashion fads and can be used as a basis for data-driven style transfer algorithms. 
Ultimately, we believe that data-driven methods applied to large-scale historical image datasets can radically change the methodologies in which visual cultural artifacts are employed in humanities research.

\section*{Acknowledgments}
The authors would like to thank Bharath Hariharan, Carl Doersch and Evan Shelhamer for their insightful comments.
This material is based upon work supported in part by the NSF Graduate Research Fellowship to Shiry Ginosar, ONR MURI N000141010934 and an NVIDIA hardware grant.

\ifCLASSOPTIONcaptionsoff
  \newpage
\fi

\bibliographystyle{IEEEtran}
\bibliography{jrnlbib}

\end{document}